\documentclass[twocolumn,10pt,journal,twoside]{IEEEtran}
\usepackage[cmex10]{amsmath}
\usepackage{amssymb}
\usepackage{amsfonts}
\usepackage{amsthm}
\usepackage{algorithm}
\usepackage{algpseudocode}
\usepackage{graphicx}
\usepackage{epsfig}
\usepackage{cite}
\usepackage{tensor}
\usepackage[caption=false,font=footnotesize]{subfig}
\usepackage{color}
\usepackage{makecell}
\usepackage{float}
\usepackage{comment}
\usepackage{subfig}
\usepackage[dvipsnames]{xcolor}
\usepackage{algpseudocode}
\usepackage{booktabs}

\usepackage{hyperref}
\hypersetup{hypertex=true,
	colorlinks=true,
	linkcolor=blue,
	anchorcolor=blue,
	citecolor=blue}

\graphicspath{{figures/}}
\DeclareGraphicsExtensions{.eps}
\theoremstyle{definition}

\newcommand\T{{\hspace{-0pt}\intercal}}

\begin{document}

\title{
A Shape Detection Framework for Deformation Objects Using Clustering Algorithms
}

\author{
Fangqing Chen $^*$\\
University of Toronto

\thanks{
Copyright may be transferred without notice, after which this version
may no longer be accessible.
}
\thanks{
$^*$ Corresponding Author.
}

}

\bstctlcite{IEEEexample:BSTcontrol}

\maketitle

\begin{abstract}
This paper uses clustering algorithms to introduce a shape framework for deformable objects.
Until now, the shape detection of the deformable objects has faced several challenges:
1) unable to form a unified framework for multiple shapes;
2) the calculation burden as a large number of calculations;
3) the inability to solve the 3D point-cloud case.
A novel shape detection framework for deformable objects is presented in this paper, which only uses the input 2D-pixel data of the objects without any artificial markers.
The proposed detection approach runs in a highly real-time manner.
For the definitions of the shapes of the deformable objects, three shape configurations are used to describe the outlines of the objects, i.e., centerline, contour, and surface.
In addition, for the obtaining of the 3D shape, 
Different from the traditional 3D point cloud processing method, this article uses a one-to-one mapping method between 2D-pixel points and 3D shape points. Therefore, this guarantees a one-to-one correspondence between 2D and 3D shape points.
Hence, the proposed approach can enhance the autonomous capability to detect the shape of deformable objects.
Detailed experimental results are conducted within the centerline configuration to evaluate the effectiveness of the proposed shape detection framework.
\end{abstract}

\IEEEpeerreviewmaketitle

\section{Introduction}\label{section1}
Deformable object manipulation (DOM) has always been a key issue that needs to be solved in many industries, such as 
industrial processing \cite{arents2022smart},
medical surgery \cite{niu2020development}, 
furniture services \cite{}, and 
food package \cite{kartika2020service}. 
However, although great progress has been made in deformable object recognition, it is still difficult to form a unified shape detection framework \cite{qi2022model}. 
One of the biggest challenges in detecting deformable objects is the complex shape outlines of objects occurring in contact with the robot during the practical application.

As for the issue of DOM, the most critical step is to do the initial shape detection \cite{meruliya2015image}.
The core idea is that the image information collected by the camera is processed into the shape characteristics of the DLO and inputted into the shape servo control system as a feedback signal \cite{qi2022towards}.
Therefore, the effect of shape detection is a key factor in determining the performance of the shape servo system and plays an extremely important role in the shape servo control process \cite{wang2009vibration}.
Meanwhile, since deformable objects have complex shape outlines, achieving a unified and accurate shape detection framework with flexible objects \cite{das2011autonomous}.
Especially in practice, the shape of the deformable objects will change with contact with the robot's effector; in this term, the soft objects are different from their rigid counterparts \cite{huang2021non}.

To the best of our knowledge, this is the first attempt to design a shape-detection framework for deformable objects using clustering algorithms \cite{nagpal2011comparative,aggarwal2018survey,joseph2019survey}, which helps to guide the robot to complete the manipulation in a low-dimensional feature space.
Detailed experiments are conducted on the centerline configuration with a large amount of shape data, and our model is compared against several state-of-the-art clustering algorithms. 
The results demonstrate that the proposed shape detection framework not only outperforms the existing methods in terms of detection accuracy but also exhibits greater real-time performance for meeting the practical requirements, which is more important in real industrial production \cite{braumann2015adaptive}.
Moreover, this paper proposes a shape detection framework based on a clustering algorithm, which will help subsequent researchers design new shape detection algorithms.

The key contributions of this paper are three-fold:
\begin{itemize}
\item 
\textbf{Construction of the Unified Shape Detection:}
This paper forms a unified detection framework for shape detection of deformable objects. The core idea is to use clustering algorithms to extract shapes.
This novel approach does not need artificial markers and can enhance real-time performance.

\item 
\textbf{Construction of the Unified Shape Description:}
Three shapes are used to describe the shape outlines of the deformable objects.
For most tasks, these three configurations are sufficient to meet actual needs.

\item 
\textbf{Completion of the Detection Processing}
Based on the design of the shape collection framework, this article also designs respective shape sorting algorithms.
The final shapes obtained in this way are a fixed number, equidistantly spaced, and ordered, which meets the prerequisites for subsequent manipulation.
\end{itemize}



\begin{figure*}[htbp]
\centering
\subfloat[Centerline]{\includegraphics[scale=0.54]{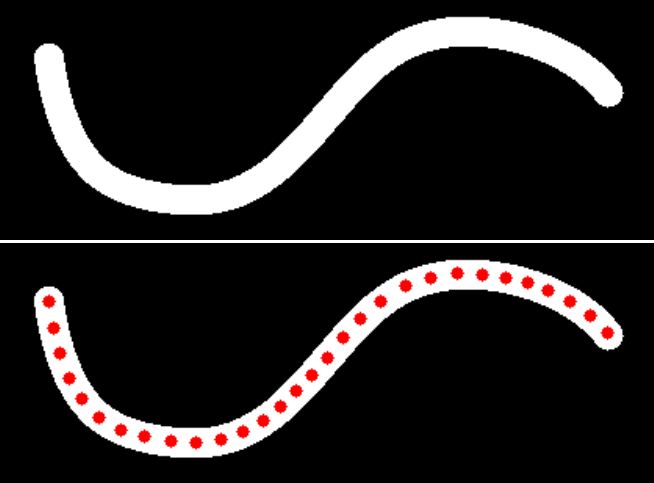}}
\subfloat[Contour]{\includegraphics[scale=0.54]{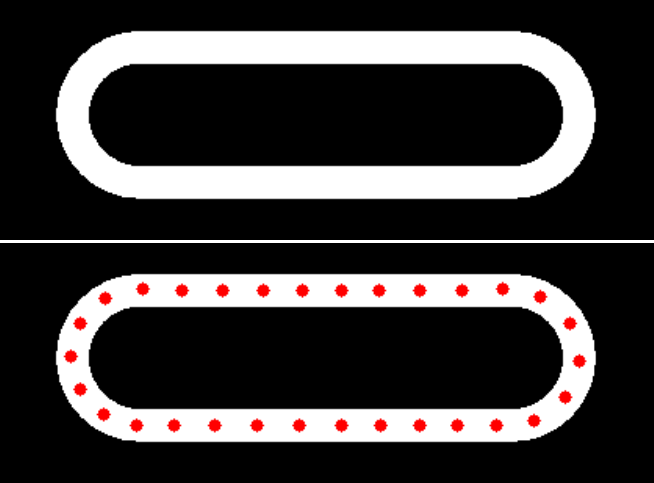}}
\subfloat[Surface]{\includegraphics[scale=0.54]{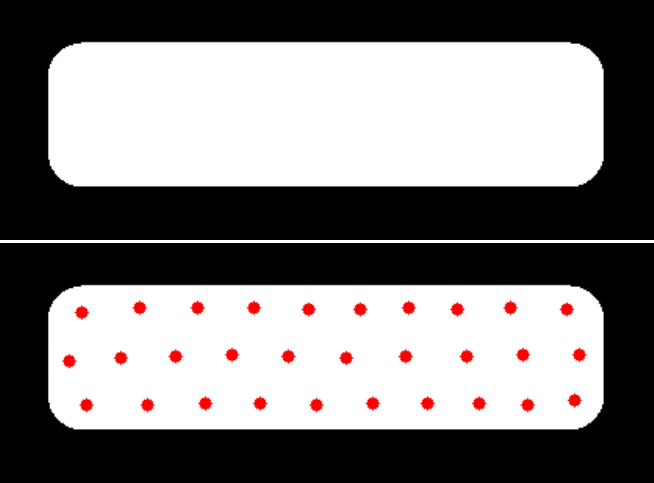}}
\caption{
Schematic diagram of shape detection based on SOM.
}
\label{fig5}
\end{figure*}




\section{Related Works}
In this section, we review the existing literature relevant to our research.
This discussion is divided into two aspects, i.e., shape representation and performance regulation.

\subsection{Shape Dection}
For manipulation, the first step is to construct the overall shape detection for the deformable objects \cite{aloi2019estimating,ma2022active,fereidoon2012manipulation,mousavi2021multifunctional}.

Points in 2D/3D space collected by color cameras or depth cameras are usually called data points. 
The simplest local shape feature (LSF) uses data points as shape features of deformable linear objects. 
Because no matter from the perspective of geometry or image processing, features based on data points are the simplest features that can be extracted.
For example, feature points \cite{vincent2005detecting,vzivkovic2004improving,qin2014image}, 
hold points \cite{petersson1999hole,gai2019research,ahmed2023vortex}, and other artificial markers \cite{ortiz2021smart,sepulveda2019automatic}.
However, shape features based on data points have a strong manual intervention, high dimensions, and low robustness to sensor measurement errors \cite{prasad2000geometric}, which can easily lead to inaccurate DLO shape description and affect system control accuracy \cite{kasim2017batik}.
Therefore, \cite{kroger2012simple} uses the ball's center of mass to represent the shape change of the middle area of the ball, which improves the anti-interference of data point features. 
Other types of LSFs, such as lines, ellipses, etc., and hybrid features are also used in some servo tasks.
However, the above techniques usually offer a high-dimensional feature vector, which may cause the calculation burden of the system or even the manipulation failure \cite{stylianou2003shape}.
There is still a lack of a shape detection framework that can describe the overall deformation state of DLO. 
Especially when facing complex manipulation tasks, the shape of DLO is ever-changing. 
Meanwhile, how to overcome the problems of illumination, contrast, environmental noise, and object surface texture in the real environment? The impact is also crucial.

\section{Clustering-based Shape Detection}
For the shape detection of the deformable objects, three shape configurations (centerline, contour, surface) are used to describe the shape of DLO during manipulation.
\begin{enumerate}
\item 
Centerline focuses on slender objects;

\item 
Contour is suitable for manipulation needs that only care about changes in the boundary of the object;

\item 
Surface is suitable for manipulation needs for the overall deformation of the object;
\end{enumerate}

These three shape configurations are sufficient to describe most shape changes.
Although many works have proposed detection algorithms for three shape configurations, the following problems still exist:
\begin{enumerate}
\item 
Fixed shape point collection cannot be guaranteed. 
For example, \emph{OpenCV/thinning} and \emph{OpenCV/findcontour} cannot guarantee fixed shape sampling. However, fixed sampling is a very important prerequisite for feature extraction;

\item 
It is impossible to detect three configurations at the same time. 
For a control environment with multiple shapes and configurations, this will increase the complexity of the system (multiple detection algorithms run simultaneously) and reduce the stability of the system;

\item 
The point cloud processing method is relatively slow for large data sets, and the point cloud measurement results are greatly affected by the external environment (illumination, contrast, brightness);

\item 
Artificial marker points need to be placed, but manual marker point placement is difficult to achieve in most cases.
\end{enumerate}

In order to solve the above problems, a shape detection method that can detect three shape configurations simultaneously was designed using SOM \cite{kohonen1996som_pak}.
SOM is a type of unsupervised learning neural network that does not need to output a data set and can perform self-organizing training using only input data. 
It can automatically find the inherent laws and essential attributes in sample data.

The core idea of the shape detection method designed in this section is to use SOM to cluster the original shape data (2D pixels or 3D point clouds) measured by the camera into the main curve of DLO/a specific number of clustering points on the surface. Similar points form different shape configurations of DLO.
Fig. \ref{fig5} shows a schematic diagram of shape detection based on SOM.
The white area of each sub-image represents the binary area (clustering area) of the target object. 
In contrast, the red dots represent the three different shape structures corresponding to the center line, contour, and surface of the shape points (clustering points) obtained by SOM clustering.
The prerequisite for realizing SOM-based shape detection is to obtain binary images, including DLO, through image processing algorithms.
Unlike conventional shape detection methods based on artificial marking points, the shape detection method presented in this article does not require the placement of artificial marking points. 
It uses the object's color to perform feature segmentation to segment the object's shape.
The image background is covered with a white curtain to simplify the shape detection algorithm.
The specific image processing process is as follows:
\begin{enumerate}
\item 
\textbf{Detection of ROI.}
Generally, marking points (for example, ArUco Marker) are placed around the platform to determine the ROI.

\item 
\textbf{Binary Image Processing.}
By setting an appropriate color threshold, convert the RGB color gamut of the ROI area to the HSV color gamut, and use mask extraction combined with a denoising function (open processing or closed processing) to smooth the image to reduce the impact of environmental noise and obtain a global binary image, set the DLO part to white and the background to black, further filter out falsely detected pixels by setting a reasonable area threshold (assuming that in the manipulation environment, the DLO part has the largest area), and obtain the white color in the binary image.
The area is the original pixel data containing DLO, which represents the pixel coordinates of each point in the image frame.
The original data is given by
\begin{equation}
\label{eq1}
\bar{\mathbf{a}} = {{\left[ {\mathbf{a}_1^\T, \dots ,\mathbf{a}_M^\T} \right ] }^\T} \in \mathbb{R}^{2M}, \ \ \ \
\mathbf{a}_i \in \mathbb{R}^2
\end{equation}

\item 
\textbf{SOM-based Shape Detection.}
Input $\bar{\mathbf{a}}$ into SOM and specify the number of clustering points $N$ to get the shape of the fixed number of points given in \eqref{eq2} represents the pixel coordinates of each cluster point in the image frame.
\begin{equation}
\label{eq2}
\bar{\mathbf{b}} = {{\left[ {\mathbf{ b}_1^\T, \dots,\mathbf{b}_N^\T} \right] }^\T} \in \mathbb{R}^{2N}, N \ll M,  \ \ \ \ 
\mathbf{b} _i \in \mathbb{R}^2
\end{equation}

\item 
\textbf{Shape Sorting Processing.}
Since $\bar{\mathbf{b}}$ is not ordered, $\bar{\mathbf{b}}$ needs to be sorted. 
In order to facilitate subsequent feature extraction algorithms, three different shape configurations have different orders.
\begin{itemize}
\item Centerline sorting.
Select the endpoint as the starting point and arrange it along the centerline axis.

\item Contour sorting.
Choose a starting point and follow the outline clockwise.

\item Surface sorting.
Arranged from top to bottom and from left to right
\end{itemize}

\item 
\textbf{3D Shape Calculation.}
Using the sorted $\bar{\mathbf{b}}$, the corresponding 3D shape can be obtained by calling the depth camera processing library $\bar{\mathbf{c}} = {{\left[ {\mathbf{c}_1 ^\T, \dots ,\mathbf{c}_N^\T} \right] }^\T} \in \mathbb{R}^{3N}, \mathbf{c}_i \in \mathbb{R} ^3$ represents the 3D coordinates in the camera coordinate system corresponding to each clustered pixel point $\mathbf{b}_i$.
At this time, $\bar{\mathbf{c}}$ is fixed sampling, ordered arrangement, and equidistant distribution.
Since 2D pixels and 3D coordinates have a one-to-one correspondence in the depth camera, this method of obtaining 3D shapes is more robust to measurement noise. 
The proposed SOM-based shape detection algorithm is simpler and more effective than traditional point cloud processing.
\end{enumerate}

\begin{figure}[htbp]
\centering
\subfloat[GMM]{\includegraphics[scale=0.70]{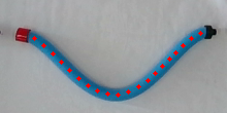}}
\hspace{0.001cm}
\subfloat[FCM]{\includegraphics[scale=0.70]{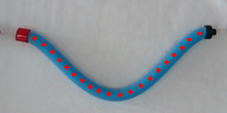}}

\subfloat[KMA]{\includegraphics[scale=0.70]{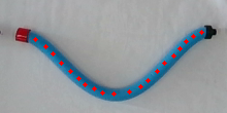}}
\hspace{0.001cm}
\subfloat[SOM]{\includegraphics[scale=0.70]{figures/fig4c.png}}
\caption{
Comparison of shape extraction effects of four clustering algorithms within centerline configuration.
}
\label{fig4}
\end{figure}

\section{Experiment Results}
This section studies the effectiveness of the proposed SOM-based shape detection algorithm.
In order to verify the proposed SOM-based shape detection algorithm, D455 was first used to shoot three different shapes of DLO.
One thousand sample data were collected for each shape configuration, and D455 used the default internal reference.
The number of shape points corresponding to each shape configuration is $N=20, N=32, N=32$.
In addition to calculation time (Time), the clustering algorithm must have high intra-class and low inter-class similarities when evaluating shape detection algorithms. 
This is reflected in shape detection and is manifested as a uniform distribution of extracted shapes. 
Each adjacent point is equidistant and covers the original geometric shape relatively comprehensively.
To this end, two evaluation indicators are considered to measure the shape extraction accuracy of the clustering algorithm
\begin{itemize}
\item 
Silhouette-Coefficient index (SC) \cite{aranganayagi2007clustering}.
The SC index value range is [-1,1]. The closer the samples of the same category are (high intra-class similarity), the farther the samples of different categories are (low inter-class similarity), the higher the SC index, which means clustering. The accuracy of the algorithm is better.

\item 
Calinski-Harabasz (CH) \cite{maulik2002performance}.
It measures the closeness within a class by calculating the sum of the squared distances between each point in the class and the center of the class. It measures the separation of the data set by calculating the sum of the squared distances between each type of center point and the center point of the data set.
The CH index is obtained from the ratio of separation and compactness.
Therefore, the larger the CH, the closer the class itself is (high intra-class similarity), and the more dispersed the classes are (low inter-class similarity), the better the clustering result.
\end{itemize}

Fig. \ref{fig4} shows the SOM and Gaussian Mixture Model (GMM)\cite{reynolds2009gaussian},
Fuzzy C-Means algorithm (FCM)\cite{bezdek1984fcm},
K-Means Algorithm (KMA)\cite{hartigan1979algorithm} of the extraction of the 2D centerline shape.
Since it is to verify the shape extraction capability of the clustering algorithm, there is no need to sort the shapes here.
It can be seen from Fig. \ref{fig4} that all four clustering algorithms can extract the centerline shape configuration well, which proves the effectiveness of applying the clustering algorithm to shape extraction.
The shape distribution extracted by KMA is not very uniform. 
It is densely distributed in some areas, while in the middle part it is sparsely sampled, so the shape clustering effect of KMA is not very good.
Intuitively, GMM has the best extraction effect (the distribution of extraction points is evenly arranged), SOM and FCM have similar effects (with slight differences in different shapes and configurations). In contrast, KMA has poor extraction effects on the three shapes and configurations.

\begin{table}[!ht]
\centering
\caption{Performance indices among four clustering algorithms}
\begin{tabular}{ccccc}
\toprule[1.5pt]
~ & SC & CH & Time (s)  \\
\midrule[1pt]
GMM & 0.342 & 47593 & 0.208    \\
FCM & 0.268 & 39072 & 0.187  \\ 
KMA & 0.254 & 34874 & 0.126   \\ 
SOM & 0.279 & 42081 & 0.063 \\
\bottomrule[1.5pt]  
\end{tabular}
\label{table1}
\end{table}

Table \ref{table1} compares performance indicators of four clustering methods when extracting three shape configurations. 
SC, CH, and Time are the average values of 1000 data samples calculated in each shape configuration.
It can be seen from table \ref{table1} that the extraction effect of GMM is the best. SOM is better when extracting the center line, and FCM is better when extracting the surface.
KMA has the lowest SC and CH indicators among the four clustering algorithms.
Although GMM has the best shape extraction effect, because GMM has a large number of iterative operations, GMM takes the longest time. Even when extracting the center line, the processing speed is only 5Hz. This high time consumption is not feasible in practical applications. Accepted.
Therefore, GMM is not suitable in environments with high real-time requirements. On the contrary, in static environments, GMM can achieve good results.
Similarly, since FCM has a large number of membership matrix calculations, the calculation time is long, and the processing speed of FCM is only about 7Hz.
Because SOM only updates local weights, its calculation time is slightly faster than KMA's.
Since the control frequency in shape servo is around 10Hz-15Hz, GMM and FCM are unsuitable for application in shape servo. However, the sampling frequency of SOM can reach 14Hz, so it can provide real-time shape detection feedback while meeting the control frequency.

The above experimental analysis proves the effectiveness of the shape detection algorithm in this paper and proves the feasibility of using the clustering algorithm as the core of shape detection.
Meanwhile, the proposed shape detection algorithm ensures the stability of shape detection and high efficiency in practical applications.

\appendices
\ifCLASSOPTIONcaptionsoff
  \newpage
\fi

\bibliography{biblio.bib}
\bibliographystyle{IEEEtran}
\end{document}